\documentclass{article}

\usepackage{PRIMEarxiv}

\usepackage[utf8]{inputenc} 
\usepackage[T1]{fontenc}    
\usepackage{hyperref}       
\usepackage{url}            
\usepackage{booktabs}       
\usepackage{amsfonts}       
\usepackage{nicefrac}       
\usepackage{microtype}      
\usepackage{lipsum}
\usepackage{fancyhdr}       
\usepackage{graphicx}       
\graphicspath{{media/}}     

\usepackage{amsmath}
\usepackage{algorithm}
\usepackage{algorithmic}
\usepackage{amsfonts}
\usepackage{tcolorbox}
\usepackage{tcolorbox}
\usepackage{listings}
\usepackage{xcolor}
\usepackage{geometry}
\usepackage{graphicx}
\usepackage{subfigure}
\usepackage{subcaption}
\usepackage{soul}
\usepackage{ulem}
\usepackage{enumerate}
\usepackage{enumitem}
\usepackage{float}
\usepackage{balance}
\usepackage{xspace}
\usepackage{balance}
\usepackage{wrapfig}
\usepackage{pifont}

\newcommand{\cmark}{\ding{51}}%
\newcommand{\xmark}{\ding{55}}%

\newcommand{\lcquad}{LC-QuAD 2.0\xspace}
\newcommand{\approach}{Q-NL Verifier\xspace}

\newtcolorbox{translatorbox}{
    width=\columnwidth,
    boxrule=1pt,
    arc=4pt,
    auto outer arc,
    left=6pt,
    right=6pt,
    top=6pt,
    bottom=6pt,
    fontupper=\small,
    title=Translation Prompt,
}

\definecolor{border}{HTML}{409F9F}

\newtcolorbox{answerbox}{
    width=\columnwidth,
    boxrule=1pt,
    arc=4pt,
    auto outer arc,
    left=6pt,
    right=6pt,
    top=6pt,
    bottom=6pt,
    fontupper=\small,
    title=LLM Answer,
    colback={border!5}, 
    colframe={border}, 
}

\newtcolorbox{reflectionbox}{
    width=\columnwidth,
    boxrule=1pt,
    arc=4pt,
    auto outer arc,
    left=6pt,
    right=6pt,
    top=6pt,
    bottom=6pt,
    fontupper=\small,
    title=Reflection Prompt,
}

\newtcolorbox{wrongbox}{
    width=\columnwidth,
    boxrule=1pt,
    arc=4pt,
    auto outer arc,
    left=6pt,
    right=6pt,
    top=6pt,
    bottom=6pt,
    fontupper=\small,
    title=Incorrect Translation,
    colback={red!5}, 
    colframe={red!40} 
}
\newtcolorbox{wrongbox2}{
    width=\columnwidth,
    boxrule=1pt,
    arc=4pt,
    auto outer arc,
    left=6pt,
    right=6pt,
    top=6pt,
    bottom=6pt,
    fontupper=\small,
    title=Incorrect Translation by Gemini - reification,
    colback={red!5}, 
    colframe={red!40} 
}

\newtcolorbox{wrongbox_verifier}{
    width=\columnwidth,
    boxrule=1pt,
    arc=4pt,
    auto outer arc,
    left=6pt,
    right=6pt,
    top=6pt,
    bottom=6pt,
    fontupper=\small,
    title=Wrong classification of the verifier,
    colback={red!5}, 
    colframe={red!40} 
}

\newtcolorbox{correctbox}{
    width=\columnwidth,
    boxrule=1pt,
    arc=4pt,
    auto outer arc,
    left=6pt,
    right=6pt,
    top=6pt,
    bottom=6pt,
    fontupper=\small,
    title=Correct Translation,
    colback={green!5}, 
    colframe={green!40} 
}

\title{Q-NL Verifier: Leveraging Synthetic Data for Robust Knowledge Graph Question Answering}

\author{
  Tim Schwabe \\
Technical University of Munich \\
Heilbronn, Germany \\
\texttt{tim.schwabe@tum.de} \\
   \And
  Louisa Siebel \\
Technical University of Munich \\
Heilbronn, Germany \\
\texttt{louisa.siebel@tum.de} \\
   \And
  Patrik Valach \\
Technical University of Munich \\
Heilbronn, Germany \\
\texttt{patrik.valach@tum.de} \\
   \And
  Maribel Acosta \\
Technical University of Munich \\
Heilbronn, Germany \\
\texttt{maribel.acosta@tum.de} \\
}

\begin{document}
\maketitle

\begin{abstract}
Question answering (QA) requires accurately aligning user questions with structured queries—a process often limited by the scarcity of high-quality query–natural language (Q–NL) pairs. 
To overcome this, we present Q-NL Verifier, an approach to generating high-quality synthetic pairs of queries and NL translations. 
Our approach relies on large language models (LLMs) to 
generate semantically precise natural language paraphrases of structured queries.
Building on these synthetic Q–NL pairs, we introduce a learned verifier component that automatically determines whether a generated paraphrase is semantically equivalent to the original query. 
Our experiments with the well-known LC-QuAD 2.0 benchmark show that Q-NL Verifier generalizes well to paraphrases from other models and even human-authored translations. Our approach strongly aligns with human judgments across varying query complexities and outperforms existing NLP metrics in assessing semantic correctness. 
We also integrate the verifier into QA pipelines, showing that verifier-filtered synthetic data has significantly higher quality in terms of translation correctness and enhances NL to Q translation accuracy. Lastly, we release an updated version of the LC-QuAD 2.0 benchmark containing our synthetic Q-NL pairs and verifier scores, offering a new resource for robust and scalable QA.
\end{abstract}

\keywords{Knowledge Graphs \and Question Answering (QA) \and Large Language Models}

\section{Introduction}

Semi-structured information stored in knowledge graphs (KGs) is invaluable to many processes and downstream applications. However, due to its semi-structured nature and reliance on specific query languages, such as SPARQL or Cypher, accessing this information remains a challenge for users without domain expertise. To bridge this gap, knowledge graph question-answering (KGQA) systems have been developed, allowing users to interact with KGs using natural language (NL). These systems translate NL questions into structured queries, enabling seamless retrieval of relevant data from the KG. As a result, QA systems play a crucial role in making (semi-)structured data more accessible to a broader audience.

\begin{wrapfigure}{l}{0.5\textwidth} 
    \centering
    \includegraphics[width=0.48\textwidth]{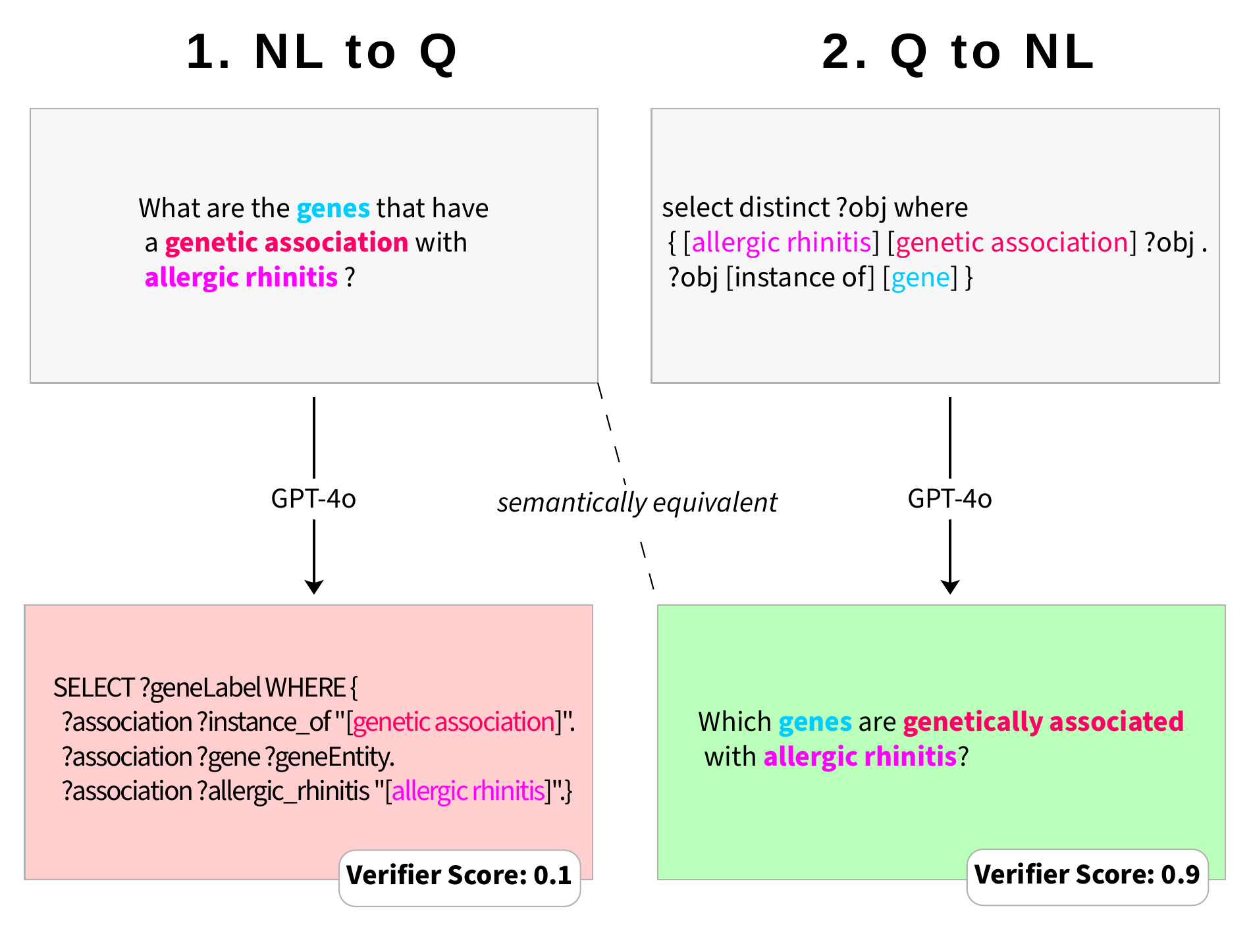} 
    \captionsetup{font=small} 
    \caption{Motivating example. 1.) GPT-4o cannot translate an NL question into a query, and our verifier predicts a low correctness score. 2.) GPT-4o can directly translate a query into natural language, and our verifier predicts correctness with high certainty.}
    \label{fig:intro}
\end{wrapfigure}

\textit{Motivation.} Building effective QA systems requires the creation of large-scale, high-quality datasets consisting of aligned query-question pairs. These datasets serve as the foundation for training machine-learning-based approaches. However, obtaining such datasets is a major bottleneck in the development of QA systems. 
Some efforts have focused on generating natural language to query (NL $\rightarrow$ Q) translations through manual approaches~\cite{Auer}.
Yet, manually curating query/NL pairs is expensive, time-consuming, and demands expert annotators to ensure semantic accuracy. The scarcity of these datasets limits the effectiveness of QA systems, particularly when dealing with complex queries involving aggregation, subqueries, or advanced filtering.
Nonetheless, recent advances in Large Language Models (LLMs) have demonstrated that these models are highly effective in generating natural language representations of structured queries (Q $\rightarrow$ NL). This observation, as shown in Figure~\ref{fig:intro} using results from GPT-4o, suggests that leveraging LLM capabilities can significantly aid in the generation of query-question datasets for QA systems.
\textit{Our Approach.} Based on these insights, we propose an approach that exploits LLM capabilities to generate (query, NL) pairs to support Knowledge Graph Question Answering (KGQA)  systems. 
Our approach begins by providing a structured query to an LLM, which then generates its natural language representation. However, these generated translations are not always accurate, necessitating techniques to assess their semantic relatedness to ensure dataset quality.
To address this issue, we propose a learned \textit{verifier}, a novel component designed to evaluate the quality of the generated translations. The verifier is capable of distinguishing between high- and low-quality translations. To implement this component, we deploy two state-of-the-art architectures based on bidirectional transformers~\cite{sentence_bert}, which are commonly used in assessing the semantic similarity of textual pairs. 
The verifier outputs a score that quantifies the semantic relatedness between structured queries and their generated natural language representations, providing an automated and reliable quality assessment mechanism.

We have implemented our approach using state-of-the-art language models of varying sizes to evaluate their impact on the quality of generated translations. Furthermore, we have compared the effectiveness of our verifier scores against standard metrics commonly used in natural language processing (NLP) for translation evaluation. Our experiments on the well-known \lcquad benchmark demonstrate that our approach yields high-quality translations, ultimately leading to improved QA systems. The results highlight the potential of LLM-generated datasets and the effectiveness of our learned Verifier in ensuring translation accuracy, enabling more scalable and robust QA systems.

\textit{Contributions and Applications.} Our main contributions are:
\begin{enumerate}
    \item We compare various frontier LLMs on their ability to translate between SPARQL and natural language paraphrases with few-shot prompting, reflection, and reasoning.
    \item We introduce a learned verifier model based on our synthetic data that can differentiate correct from incorrect translations and show that it improves over common NLP metrics for assessing query translation.
    \item We show that the synthetic data, in combination with the verifier, can be used to train QA systems.
    \item We provide an updated version of the \lcquad benchmark based on our synthetic translations and verifier scores with significantly improved quality.
\end{enumerate}

The verifier has a broad set of downstream tasks or applications: cleaning datasets for QA systems, feedback signal in training of learned QA models, during usage of QA models to catch wrong translations as well as interactive feedback for users that want to manually translate natural language questions to queries.
Lastly, we show that the synthetic translations, in combination with the verifier, can be used as training data for QA systems.

\textit{Paper Structure.} The remainder of this paper is structured as follows. 
Section~\ref{sec:preliminaries} provides background on QA over knowledge graphs and LLMs. 
Section~\ref{sec:related_work} details related work, and Section 4 presents our approach \approach. 
Section 5 presents our experimental study. In Section 6, we describe a newly released version of the \lcquad benchmark generated with \approach. Finally, Section 7 concludes our key findings and future research.
\section{Preliminaries}
\label{sec:preliminaries}
\subsection{Question Answering (QA) Systems}
A QA system aims to retrieve the correct answer to a user’s (natural language) question. This paper focuses on QA systems over knowledge graphs (KGs). A KG is composed of entities and relationships represented as a directed graph $G = (E, R)$
where $E$ is the set of entities and $R$ is the set of relations linking those entities. Questions over Knowledge Graphs are formulated using query languages like SPARQL. A given question in natural language $t$ can be reformulated to a semantically equivalent query $q$. The goal of a QA system is to perform the correct conversion from $t$ to $q$ in order to retrieve factual answers based on the underlying KG.

\subsection{Large Language Models (LLMs)}
An LLM is a neural network–based architecture (most often a decoder-only Transformer) trained on massive text corpora. Given a prompt $P$, the LLM autoregressively predicts the next words, conditioning on both the prompt and all words generated so far, thereby generating a response $R$. By iterating this process word by word, the model produces coherent text that can be used in different applications, including text completion, summarization, question-answering, programming, and more.

More recently, so-called \textit{Large Reasoning models} have been introduced. Instead of directly returning an answer, they first produce a long \textit{chain of thought}, akin to an inner monologue. Within this monologue, the model reasons about the best answer to the prompt, reevaluates its proposed answer and backtracks if necessary. This has been shown to significantly improve performance on various benchmarks~\cite{o1_eval}.

\section{Related Work}
\label{sec:related_work}
The problem of generating queries and their corresponding natural language questions has been approached from multiple perspectives. 
Earlier works proposed \textit{rule- and template-based} approaches to generate translations. 
More recent works rely on \textit{machine learning} and provide solutions either for the \textit{Q $\rightarrow$ NL translation} task, or the \textit{NL $\rightarrow$ Q translation} task. 

\paragraph{Rule- and Template-Based Translations.}
\textit{SPARQL2NL}~\cite{sparql2nl} uses a rule-based approach that first normalizes the query structure, followed by handcrafted linguistic rules to verbalize the query. The approach shows good results on simpler queries but degrades for more complex ones beyond the defined linguistic patterns. \textit{LD2NL}~\cite{ld2nl} extends \textit{SPARQL2NL} by using a bottom-up approach that uses grouping and aggregation and also enables translation of RDF triples and OWL axioms.
Regarding approaches to translating NL to SPARQL, Unger et al.~\cite{template2} propose to first extract the syntactic structure of an NL question and then inject this into SPARQL query templates. Due to its rule-based nature, this approach is also challenged by more complex query structures.
At the intersection between rule-based systems and machine learning, \textit{TeBaQA}~\cite{tebaqa} uses a trained classifier to identify to which query pattern a given question is isomorphic and then uses a rule-based approach to instantiate the query pattern. As with other solutions, this approach is restricted to the defined query patterns.
To overcome this limitation, in this work, we propose a solution that does not rely on rules or templates. 

\paragraph{Machine Learning Based NL $\rightarrow$ Q Translations}
Multiple recent approaches have proposed using LLMs to translate natural language questions to SPARQL queries. Rangel et al.~\cite{rangel} finetuned a small OpenLLaMA model for SPARQL query generation in the life science domain and extended the dataset by augmenting existing SPARQL/NL pairs. SGPT~\cite{sgpt} trains GPT-2 for NL to SPARQL translation by incorporating linguistic features through an embedding layer into the model. Yin et al.~\cite{yin} and Diallo et al.~\cite{diallo} investigate different architectures of neural machine translation for SPARQL.
These works differ from ours as they tackle the problem of generating Q/NL pairs by translating NL questions into structured queries; in this work, we tackle this problem in the opposite direction.

\paragraph{Machine Learning Based Q $\rightarrow$ NL Translations}
This line of research is the closest to our work. 
NABU~\cite{nabu} combines Graph Neural Networks and transformers to verbalize RDF triples. This removes the need for complex templates and rule-based approaches and generalizes to new queries and graphs. Unlike our approach, this approach trains a model from scratch and hence needs a large amount of labeled data (RDF triples and verbalization). 
In the context of conversational QA, Lecorve et al.~\cite{lecorve} finetune older generations of encoder-decoder transformers for the SPARQL to NL task, which work well for frequently observed query templates but struggle with complex and rare ones. 
Preevalov et al.~\cite{areWeThere} propose a multilingual approach based on modern decoder-only transformer models (GPT-4, Mistral) to verbalize SPARQL queries to non-expert users. Similar to us, they inject label information about the occurring entities into the prompt but do not use further descriptions about the entities and do not perform reflection. Further, they do not include an automatic filter for the translations. FlexKBQA~\cite{flexKBQA} similarly uses modern decoder-only models (GPT-3.5) to translate queries to natural language. FlexKBQA relies on training smaller language models with a combination of real and generated SPARQL/NL pairs. While this approach includes a verification step, it does not utilize a trained verifier specifically for SPARQL-NL similarity or incorporate reflection-based improvements in their prompting strategy.
\begin{figure*}[t!]    \includegraphics[width=\linewidth]{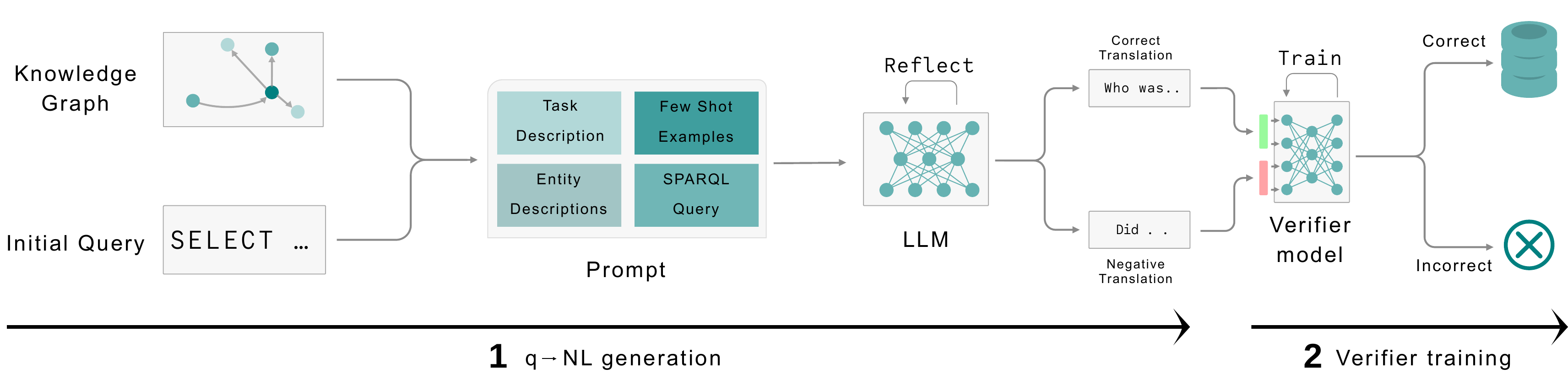} 
    \caption{Our approach: \approach. (1) Pipeline for synthetic SPARQL/NL pair generation by tackling the Q $\rightarrow$ NL task. (2) The pairs previously generated are used to train the verifier model.}
    \label{fig:workflow}
\end{figure*}
\section{Our Approach: Q-NL Verifier} 
\label{sec:q_to_nl}

We propose a learned verifier, called Q-NL Verifier, a novel component to assess the quality of pairs of queries and their natural language counterparts (Q/NL pairs) used in Knowledge Graph Question Answering (KGQA) systems. 
Figure \ref{fig:workflow} depicts our approach to building the Q-NL Verifier. 
The first part consists of generating synthetic translations by tackling the  Q $\rightarrow$ NL task (Section~\ref{sec:q_nl_generation}). 
Here, we rely on the knowledge graph (KG) and a large language model (LLM) to generate the synthetic translations, which are used to train the verifier (Section~\ref{sec:verifier}).
Lastly, the synthetic Q/NL pairs identified as correct by the verifier can be used in a wide range of downstream tasks (Section~\ref{sec:applications}). 
In the following, we provide details about all these aspects of our approach. 


\subsection{Q $\rightarrow$ NL Generation}
\label{sec:q_nl_generation}
As shown in Figure~\ref{fig:workflow}, the input for the Q $\rightarrow$ NL generation is a structured query and a knowledge graph (KG). 
In this work, we consider KGs modeled in a triple-based representation \textit{(head, relation, tail)}, such as the Resource Description Framework (RDF)~\cite{RDF}, as this is the model assumed in most QA systems~\cite{dimitrakis}. 
For such KGs, the recommended query language is SPARQL~\cite{sparql11}. 

The goal of this part of the pipeline is to generate natural language translations for SPARQL queries using LLMs (cf. Algorithm~\ref{alg:translation_sparql}). 
For this, we implement prompt construction techniques that facilitate the LLM in understanding the SPARQL query and increase the quality of the translation. 
Furthermore, in our pipeline, we use reasoning techniques (such as self-reflection) to increase the quality of the translations.    
\begin{algorithm}[h]
\caption{Q $\rightarrow$ NL Generation}
\label{alg:translation_sparql}
\textbf{Input:} SPARQL query $q$, RDF graph $G$, Language Model $M$, a set of example translations $(q_i,t_i)$
\begin{algorithmic}[1]        
    \STATE $D \gets \text{get\_descriptions}(q, G)$ \COMMENT{\textit{Descriptions of entities/relations from $G$}}
    
    \STATE $q' \gets \text{replace\_ids}(q, G)$ \COMMENT{\textit{Replace entity/predicate IRIs with labels from $G$}}

    \STATE $P \gets \text{buildPrompt}(q', D)$
    
    \STATE $t \gets M(P)$ \COMMENT{\textit{Translate query using model $M$}}
    
    \STATE $R \gets \text{buildReflectionPrompt}(q', D, T)$ \COMMENT{\textit{Prompt for reflection and improvement}}
    
    \STATE $t \gets M(R)$ \COMMENT{\textit{Reflected and improved translation}}
\end{algorithmic}
\end{algorithm}

\begin{figure}

\input{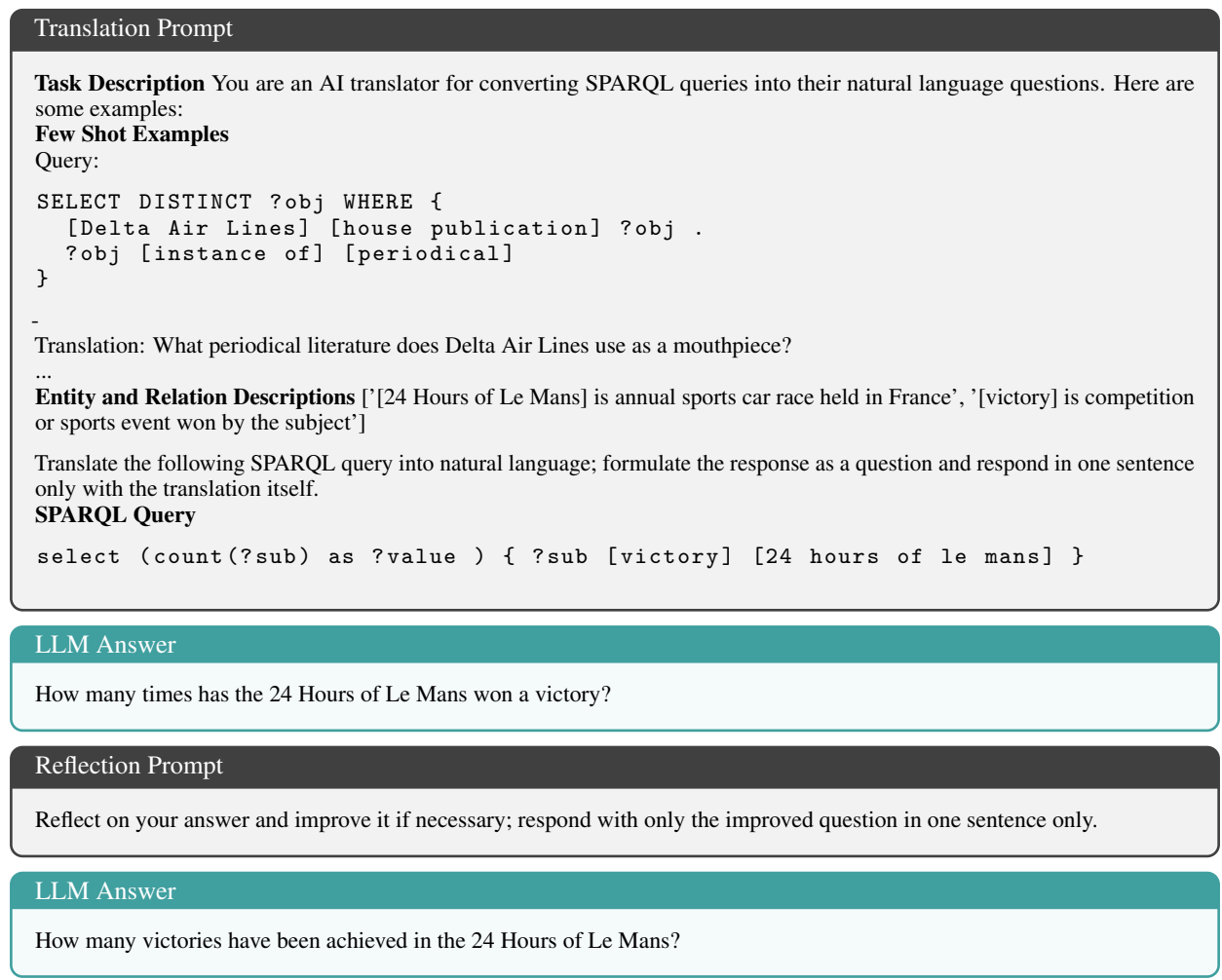}

\begin{answerbox}
How many times has the 24 Hours of Le Mans won a victory?
\end{answerbox}

\begin{reflectionbox}
Reflect on your answer and improve it if necessary; respond with only the improved question in one sentence only.
\end{reflectionbox}

\begin{answerbox}
How many victories have been achieved in the 24 Hours of Le Mans?
\end{answerbox}
\caption{Sequence of prompts and LLM answers produced by our approach during Q $\rightarrow$ NL generation. The SPARQL query is taken from the \lcquad benchmark. }
\label{fig:prompt}
\end{figure}

\subsubsection{Prompt Construction}
For every SPARQL query to be translated, our approach constructs an enriched prompt to instruct the LLM to perform the translation. 
These prompts consist of: 
(1)~a task description,  
(2)~few shot examples, 
(3)~entity and relation descriptions,  
(4)~a reformulation of the SPARQL query to be translated.

\paragraph{(1) Task Description} 
This is a fixed text at the beginning of each prompt and provides a short specification of the task to the LLM.

\paragraph{(2) Few Shot Examples}
The prompt includes $k$ examples of correct translations to show the model how to solve the provided task. 
In machine learning, this technique is called \textit{few shot} and has been shown to improve the quality of the model's output~\cite{brown}. 
In our approach, the examples are also considered fixed or the same for all the prompts. In other words, the examples do not have to be adjusted to the query to be translated and they do not even have to be from the same knowledge domain. 
This facilitates the application of our approach to different KGs, as it does not require domain-specific examples.   
In the implementation of our approach, we provide $k=4$ examples, which are based on existing queries included in the \lcquad benchmark. 

\paragraph{(3) Entity and Relation Descriptions}
This part of the prompt includes short descriptions in natural language about the entities and relations that occur in the SPARQL query. 
These descriptions enable the LLM to better understand the type and nature of the entities mentioned in the query, and how they are connected through the relations. 
Our approach retrieves these natural language descriptions automatically from the underlying KG (Algorithm~\ref{alg:translation_sparql}, line 1).
Most RDF KGs -- such as DBpedia and YAGO -- provide these descriptions using the recommended predicate \texttt{rdfs:comment} from the RDFS vocabulary~\cite{RDF}.  
Other datasets, such as Wikidata, use other predicates such as \texttt{schema:description} from the Schema.org~\cite{SchemaOrg} vocabulary to provide human-readable descriptions~\cite{WikidataGlossary}.

\paragraph{(4) (Reformulated) SPARQL Query}
This corresponds to the query to be translated by the LLM.  
In SPARQL, entities and relations mentioned in the query are identified with IRIs. 
This is because the underlying data model RDF employs IRIs as global identifiers. 
However, IRIs can be long, and in some KGs (such as Wikidata), they correspond to sequential identifiers without any meaning. For example, the entity \texttt{Marie Curie} is identified as \texttt{wd:Q6762812} in Wikidata.\footnote{This is actually an abbreviation, the full IRI of Marie Curie is \url{https://www.wikidata.org/entity/Q6762812}.}  
The inclusion of IRIs in the prompt is problematic for the LLM, as they do not carry any semantic information that can be used by the LLM to generate the translation. 
Therefore, our approach reformulates a SPARQL query $q$ by $q'$, which replaces IRIs in $q$ with their corresponding natural language labels. This is achieved by retrieving from the RDF KG, the value of the predicate \texttt{rdfs:label} for each entity and relation that occurs in $q$ (Algorithm~\ref{alg:translation_sparql}, line 2). The predicate \texttt{rdfs:label} is the de facto way to include human-readable names for resources in RDF. 




After retrieving from the knowledge graph the entity and relation descriptions $D$ as well as the SPARQL query with human-readable labels $q'$, our approach builds a prompt $P$ for translating the query $q$ (Algorithm~\ref{alg:translation_sparql}, line 3).  
Figure~\ref{fig:prompt} shows an example of the translation prompt built by our approach.

\subsubsection{Translation using LLM}\label{sec:llm_translation}
The constructed prompt $P$ is then used to instruct an LLM $M$ to generate an NL translation (Algorithm~\ref{alg:translation_sparql}, line 4). 
In the implementation of our approach, we access different LLMs through their APIs. 
The LLM produces an answer which is the synthetic translation $t$ of the query $q'$. 
In some cases, this translation is still prone to errors, e.g., hallucinations where terms in the natural language paraphrases are not mentioned in the input query, or the translation contains wrong associations between the entities. 
To minimize these errors, LLMs can perform \textit{reflection}~\cite{renze} to re-process their answer and improve them if necessary. 
This technique generally increases the model's response quality~\cite{renze}. 
Standard LLMs (llama3, Gemini, and GPT 4o) require explicit instructions to perform reflection, while \textit{reasoning models} like o1-mini already perform extensive internal reflection.
Figure~\ref{fig:prompt} shows how our approach builds a reflection prompt $R$ over such models, leading to a new, improved translation $t$.


\subsection{Verifier Training}
\label{sec:verifier}

Each synthetic translation $t$ created using the Q $\rightarrow$ NL generation is used to train the verifier model (Figure~\ref{fig:workflow}). 
The model learns to distinguish correct from wrong translations. 
We model this problem as the problem of estimating the similarity between two textual inputs ($q$ and $t$) by computing a quality score (or \textit{verifier score}): 
\begin{equation}
\textit{Verifier}_{Q,NL} = s(q,t) \in [0,1].
\end{equation}
where $s$ is the similarity between $q$ and $t$; 0 indicates no similarity, and 1 indicates full similarity. 
If the verifier score is above a threshold $\tau$, then the translation $t$ is considered correct; otherwise, it is considered incorrect.  
Our verifier requires only the SPARQL query and its synthetic translation to compute similarity, unlike existing approaches that depend on gold-standard translations. This eliminates the need for external validation data, making our approach more scalable and efficient.

To learn the verifier model, we draw inspiration from related, more general work of modeling textual similarity~\cite{sentence_bert} and apply bi- and cross-encoders. 
These are state-of-the-art architectures to build embeddings -- i.e., multi-dimensional numerical vectors --   of textual inputs (or sequences of tokens) that are able to capture the similarity between the texts effectively. 

\begin{figure}[t!]
\centering
\includegraphics[width=0.5\linewidth]{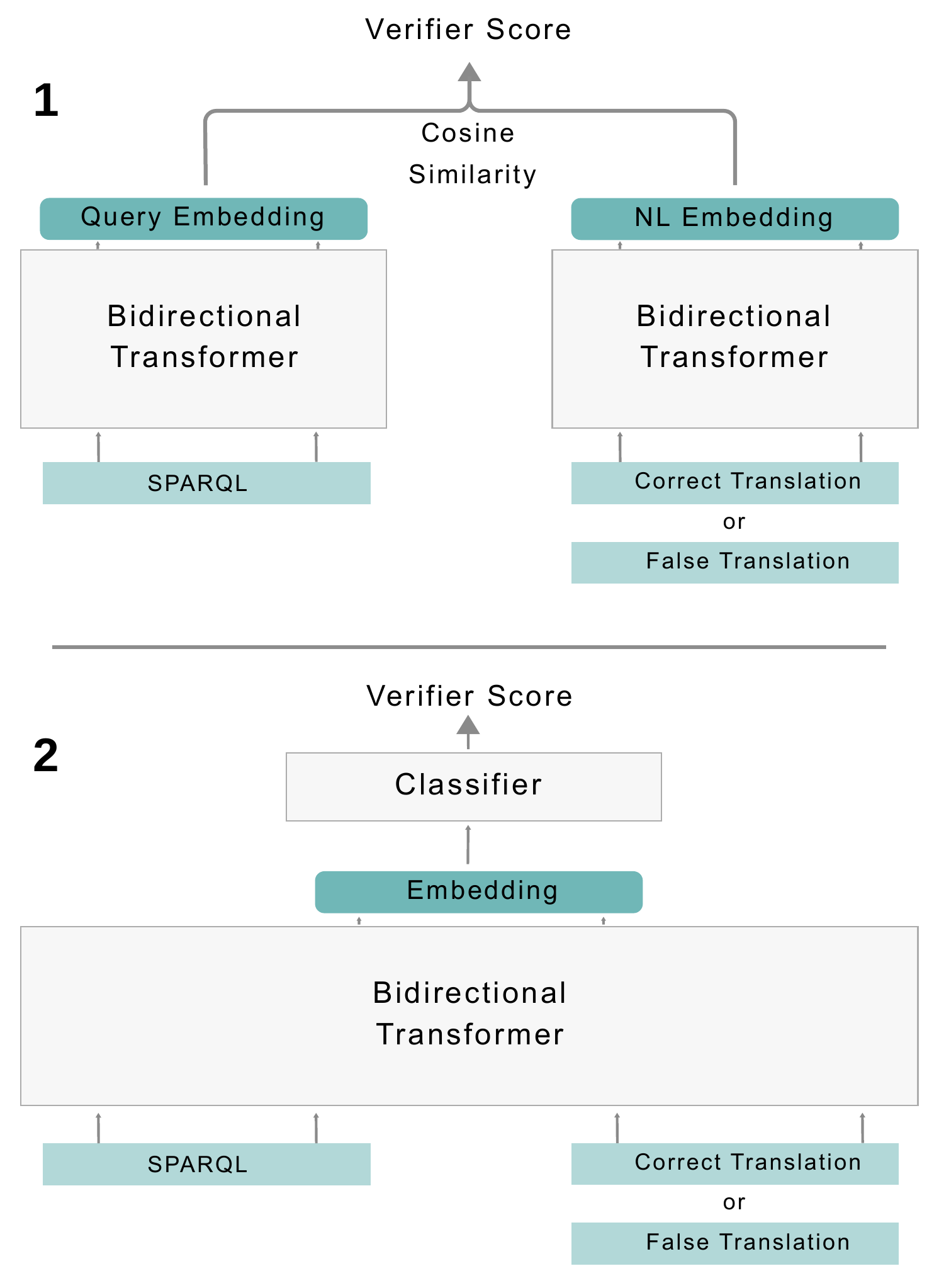} 
    \captionsetup{font=small} 
    \caption{Verifier models for measuring the similarity between a query and its natural language translation. (1) A Bi-Encoder processes the query and translation independently into latent vectors, and their similarity is measured using cosine similarity. (2) A Cross-Encoder jointly processes the query and translation within one model and predicts similarity as a scalar value between 0 and 1 via a classifier.}
    \label{fig:verifier}
\end{figure}

\smallbreak
\noindent
\textbf{Bi- and Cross Encoders.}
Bi- and cross-encoder architectures capture the semantic similarity between two textual inputs, such as determining if two sentences share the same meaning or if a text answers a specific question~\cite{sentence_bert}. 
Bi- and cross-encoder models are types of transformer architectures that use bidirectional attention~\cite{sentence_bert}.  
To build the embeddings, bidirectional attention transformers process all the tokens at once. This allows each token within an input sequence to attend to all other tokens, allowing maximal exchange of contextual information across the entire sequence. 
In contrast, the traditional causal masked attention mechanism (also called autoregressive next-token prediction) processes the tokens from left to right, focusing on tokens preceding the current position. 
This limits the exchange of contextual information across the entire sequence in causal masked attention.  
For this reason, in this work, we use bidirectional attention, which yields richer semantic representations, which is important for semantic similarity.




The architectures of bi- and cross-encoders are different, as depicted in Figure~\ref{fig:verifier}.
Bi-encoders process each text input independently using a shared encoder model, generating two embedding vectors (768 dimensions each). In our case, one embedding for the SPARQL query $q$, and another one for the NL translation $t$.  
These embeddings are compared using a similarity metric to compute $s$.
In our work, we use the cosine similarity --   
defined as the normalized dot product of the two vectors or embeddings --  as it captures whether the embeddings have similar orientations. 
In sub-symbolic spaces, embeddings of tokens used in similar contexts in the input text tend to have similar directions. 
I.e., A high cosine similarity value indicates a high semantic similarity between the input texts. 

Unlike bi-encoders, cross-encoders use a single model to process both inputs jointly by concatenating them. This approach captures the semantic similarity between the input texts during this joint processing step.
The obtained embedding of both inputs is fed into a classifier to directly predict $s$. 
Following the literature, our classifier is also implemented as a one-layer neural network,  which transforms the embedding from the encoder into a value between $0$ and $1$ that represents the similarity $s$ between the input texts.

\smallbreak
\noindent
\textbf{Training.}
Training both bi-encoders and cross-encoders typically relies on positive and negative pairs of inputs. 
Positive pairs consist of semantically similar texts and negative pairs are dissimilar~\cite{sentence_bert}. 
The reason is that only providing positive pairs constitutes a trivial learning problem, where perfect accuracy is achieved by always predicting 1 (correct translation), irrespective of the input. In our case, this means the verifier models must be presented with pairs of queries with correct (positive) and wrong (negative) translations during training. 
To generate negative examples, one approach is to randomly use the translation of another query in the dataset and use it as a negative example. Yet, this does not lead to a powerful model since the task can easily be solved by checking if the named entities and predicates overlap in $q$ and $t$. 
Therefore, the goal is to generate \textit{hard negative} examples, where $t$ and $q$ mention the same terms, but the translation is still incorrect. 
To generate hard negative examples, we implemented the following approach. Following the workflow to generate a (correct) translation detailed Section~\ref{sec:q_nl_generation}, we ask the model to generate a very similar translation $t_{neg}$ that, however, has a distinctly different semantic meaning from $q$ and $t$.

Using the resulting dataset of positive and negative pairs, the models are optimized to maximize the similarity for positive pairs while minimizing it for negative pairs in a supervised way. For bi-encoders, this is achieved by minimizing the contrastive loss over the embedding vectors of $q$ and $t$/$t_{neg}$. It rewards the model when positive pairs have high cosine similarity and penalizes it when negative pairs have high similarity. For cross-encoders, which jointly process both $q$ and $t$/$t_{neg}$ in a pair, we recast the problem as a binary classification task. Here, the model directly predicts whether a given pair is correct or incorrect. We use binary cross-entropy loss to train the model; this loss measures the difference between the model’s predicted probability of the pair being similar and the actual similarity (0 or 1).
After training, the models can be used to verify the semantic similarity of a query and its translation for new pairs. Despite being trained on synthetic data generated by an LLM, we show in Section~\ref{sec:experiments} that this discriminative power is generalized to pairs from other LLMs and human-generated translations.

\subsection{Downstream Tasks}
\label{sec:applications}

While the Q-NL Verifier has been initially devised to tackle the Q $\rightarrow$ NL task, it can be applied to other QA-related problems. 
In the following, we briefly discuss further downstream tasks and applications of our proposed approach. 

\paragraph{Semantic Similarity Metric} 
At the core of our Q-NL Verifier is the \textit{Verifier$_{Q, NL}$} score, which estimates the semantic similarity between a query $Q$ and its natural language translation $NL$.  
Therefore, our verifier score can be used as an alternative metric to commonly used NLP metrics.  
In our experiments (cf. Section~\ref{sec:metric}), we empirically compare the effectiveness of our \textit{Verifier$_{Q, NL}$} score to standard NLP metrics such as BLEU, Rouge, and BERTScore.  

\paragraph{NL $\rightarrow$ Q Translations with LLMs.} 
The synthetically generated pairs generated by our approach and the verifier can assist LLMs in translating natural language questions into SPARQL queries. 
This task is central to QA systems. 
The idea is to fine-tune the LLMs, using \textit{supervised finetuning}, where the model is directly trained to predict the correct SPARQL query token by token using the synthetically generated pairs.
To test this, we conducted experiments (cf. Section~\ref{sec:nl_to_q}) where we evaluated (i) the original model (LLM without finetuning), (ii) the fined-tuned model (LLM finetuned with our synthetic examples), and (iii) a model that is trained on filtered data using the verifier. 
This allows us to understand the impact of each of the NL-Q Verifier components on the NL $\rightarrow$ Q task.  



\paragraph{Feedback within QA systems}
The verifier can serve as an internal feedback mechanism within a QA system. When the system generates a query translation for a user question, the Q-NL Verifier can assess whether the translation is semantically equivalent to the input. If the verifier determines that the translation is likely incorrect or has low confidence, the system can take corrective actions such as refining the translation, attempting an alternative parsing strategy, or prompting the user for clarification. This self-assessment capability enhances the robustness of QA systems by preventing incorrect query execution.
\paragraph{User support.}
Lastly, the verifier can be employed as an assistive tool for users who manually translate natural language questions into SPARQL queries. Writing SPARQL queries requires familiarity with the underlying knowledge graph schema, making it challenging for many users, especially those without extensive expertise in semantic web technologies. By providing real-time feedback, the verifier can indicate whether a user-constructed query accurately reflects the intent of their natural language question. This can be integrated into query-building interfaces, where the system suggests corrections or highlights potential mismatches between the intended meaning and the generated query. 
\section{Experimental Study}
\label{sec:experiments}
In the experiments, we first investigate research questions associated with the effectiveness of \approach: 
\begin{enumerate}[label=\textbf{(RQ\arabic*)}]
\item Which of the studied modern LLMs is the most effective for Q$\rightarrow$NL translation to train the verifier? 
\item Which verifier architecture (cross-encoder or bi-encoder) can better distinguish between correct and incorrect translations? 
\item Does the verifier improve the overall quality of Q/NL datasets? 
\end{enumerate}

Next, we investigate research questions related to the application of \approach as described in Section~\ref{sec:applications}: 
\begin{enumerate}[label=\textbf{(RQ\arabic*)}, start=4]
\item Can \approach be used as a replacement for NLP metrics for query translation? 
\item  Does the verifier improve NL $\rightarrow$ Q translations with LLMs?  
\end{enumerate}

\subsection{Experimental Setup}\label{sec:experimental_setup}
\noindent
\textbf{Dataset.}
We use the well-known benchmark \lcquad~\cite{lcquad}. The dataset has a total of $24,000$ questions, a natural language verbalization of those, and the corresponding SPARQL queries. The dataset has queries for both Wikidata and DBpedia 2018, but we focus solely on the Wikidata queries. The natural language verbalizations of the queries were generated using Amazon Mechanical Turk. The dataset contains various queries, including single facts, multi-joins, blank nodes, aggregations using \texttt{COUNT}, filtering of values with \texttt{FILTER}, etc. 
\smallbreak
\noindent
\textbf{Language Models.}
We tested several different LLMs, from small open-source models to large proprietary frontier models.

\begin{itemize}[leftmargin=5mm]
\item \textit{Llama 3 7B. }
Released in April 2024, Llama 3 7B is an open-source language model with 7 Billion parameters, designed to natively support multilinguality, coding, reasoning, and tool use. It is part of the Llama 3 family, which includes larger variants such as 70B and 405B parameter models. Llama 3 7B demonstrates excellent performance among models of similar size, achieving strong results on common benchmarks across coding, question answering, and other tasks. Its architecture is a standard dense Transformer, with improvements stemming from high-quality, diverse training data and increased training scale. We include Llama 3 7B to assess the translation performance of small, open-source language models.

\item \textit{Gemini 2.0 Flash. }
Introduced in December 2024, Gemini 2.0 Flash is Google's latest experimental AI model, building upon the previous Gemini 1.5 Flash. It supports multimodal inputs and outputs, including text, images, and audio.  Gemini 2.0 Flash outperforms its predecessors in various benchmarks. We use it with the Gemini API in Google AI Studio. We include this model as yet another frontier model.

\item \textit{GPT-4o. }
Released in May 2024, GPT-4o is OpenAI's multilingual and multimodal non-reasoning flagship model. GPT-4o achieves state-of-the-art results in multilingual and vision benchmarks. We use it via the OpenAI API. We include this model as representing one of the top-performing frontier models.

\item \textit{o1-mini. }
Recently published, o1-mini is a so-called Large Reasoning Model. It generates a long chain of thought to reason about the prompted task before answering. This reasoning is hidden from the user, and only the final answer is visible. Reasoning models like o1-mini have shown substantial improvements on challenging benchmarks. We include this model to evaluate the new paradigm of reasoning models. 
\end{itemize}

\noindent
\textbf{Approach Implementation.}
All experiments are implemented using Python. The LLMs are accessed using their respective APIs or run locally using Ollama~\cite{ollama}. The verifier models are implemented using \textit{SentenceTransformers}~\cite{sentence_bert}.
As the base model for our verifier model, we use \textit{ModernBERT}~\cite{modernbert}. Recently published, it is an encoder-only Transformer model trained using masked token prediction and has shown improved benchmarks compared to older generations of BERT-type models with improved efficiency. Our code is publicly available on GitHub.\footnote{\url{https://github.com/TimEricSchwabe/Q-NL-Verifier}}

\smallbreak
\noindent
\textbf{Evaluation Metrics.}
In order to assess the quality of translations of SPARQL queries and the verifier, we use automatic metrics that are commonly used in NLP to assess the semantic similarity of text pieces. However, since the given ground truth translations in the \lcquad dataset are often incorrect (see section~\ref{sec:results_q_to_nl}), we also performed extensive manual checks of the translations.

\begin{itemize}[leftmargin=5mm]
\item \textit{Syntactic metrics.} As metrics commonly used in NLP, we use \textit{BLEU}~\cite{bleu}, \textit{Rouge L}~\cite{rouge}, and \textit{Levenshtein Distance}~\cite{levenshtein} between the natural language translation (\textit{NL}) and the ground truth from the dataset (\textit{GT}). 
BLEU works by computing the overlap in N-grams between translation and ground truth, Rouge L looks at the longest common subsequence between those, and Levenshtein Distance measures how many single-character edits (insertion, deletion, substitution) are necessary to transform the translation to the ground truth. 

\item \textit{BERTScore.} While informative, the previous metrics focus on syntactic aspects and do not capture the semantic similarity between the sentences. 
To this end, we use \textit{BERTScore}~\cite{bert_score}, which measures the cosine similarity between BERT embeddings of the translation and ground truth. This metric has been shown to sensitively represent semantic similarity between pairs of sentences. For this method, we use the ModernBERT model mentioned above. We term this metric $BERTScore_{NL, GT}$.

\item \textit{Verifier Score (Our Approach).} We use the developed verifier model that measures the similarity of a query and a given translation. Since this metric only depends on a query and a translation, it does not suffer from any inaccurate reference translations. We term this metric $\textit{Verifier}_{Q, NL}$.

\item \textit{Manual Accuracy.} 
Since all metrics mentioned above only have a finite accuracy in assessing semantic similarity between translation and ground truth (in addition to the quality issues of the dataset), we also evaluated the translation of the selected queries for each model manually. For that, we investigated the original SPARQL query and the occurring entities/predicates and assessed whether the translation given by the LLM is a correct semantic verbalization of the query. We report this metric as $Acc_{manual}$, i.e., how many percent of the tested queries are correctly translated.
\end{itemize}


\begin{table}[t!]
  \caption{Accuracy (manually computed) of the model performance for SPARQL query to NL translation on 300 queries from the  \lcquad dataset. LLMs with reasoning (Reas.) capabilities are marked. }
  \label{tab:model-acc-manual}
  \centering
  \begin{tabular}{lcccc}
    \toprule
    \textbf{Model} & \textbf{Type} & \textbf{Size} & \textbf{Reas.} & \textbf{$Acc_{manual}$} \\
    \midrule
    Llama 3 7B  & Open Source& \textit{Small} &  \xmark &0.61 \\
    Gemini 2.0 Flash  & Frontier & \textit{Large} & \xmark & 0.88 \\
    o1-mini  & Frontier & \textit{Large}  & \cmark & 0.94 \\
    GPT-4o  & Frontier & \textit{Large} & \xmark& \(\textbf{0.97}\) \\
    \midrule
    \lcquad & -- & -- & -- & 0.60 \\
    \bottomrule
  \end{tabular}
\end{table}

\subsection{Results: Q $\rightarrow$ NL Translation Accuracy}
\label{sec:results_q_to_nl}

In this experiment, we assess the accuracy of the LLMs in the Q $\rightarrow$ NL Translation. The goal of this study is to identify the most performant LLM to implement the generation of synthetic Q/NL pairs in \approach. 
For this, we randomly selected $300$ queries from the \lcquad benchmark.  

\begin{figure}[t!]

\subfigure[Correct translation by GPT-4o]{
\input{correct_example}
\label{fig:ex_correct}
}


\subfigure[Incorrect translation by GPT-4o: inverted relation]{
\input{incorrect_example2}
\label{fig:ex_incorrect1}
}

\subfigure[Incorrect translation by Gemini: reified statement ignored]{
\input{incorrect_example3}
\label{fig:ex_incorrect2}
}

\caption{Examples of translations with different LLMs}
\label{fig:gpt_translation}
\end{figure}

To measure the accuracy of the LLMs, we manually inspected the SPARQL queries and their translations into natural language. 
This ensures that our analysis is not sensitive to existing reference translations or to the selection of specific metrics to compute accuracy. 
Table~\ref{tab:model-acc-manual} shows the results of the manually measured accuracy for the studied LLMs. 
While all models achieve high accuracy (above 0.50) in terms of correct translations, and there is a clear trend from small open-source models to large frontier models. 
Most notably, GPT-4o performs a correct translation in over 97 \% of the queries. 
Figure~\ref{fig:ex_correct} shows an example of one correct translation by GPT-4o. 
Existing errors in the translations are mostly the usage of inverted properties -- e.g., using \textit{father of} instead of \textit{children of } --  which can result from descriptions of predicates that are not detailed enough. 
An example of such incorrect translations is shown in Figure~\ref{fig:ex_incorrect1}.
Further, Llama 3 7B, Gemini 2.0 Flash, and o1-mini sometimes misinterpret query variables and formulate recursive statements when dealing with the reified statements in Wikidata.  
An example of such incorrect translations is shown in Figure~\ref{fig:ex_incorrect2}. 
Llama 3 7B frequently hallucinates and adds additional statements to the translation.
In terms of the best model, GPT-4o, all translations are grammatically correct and, if wrong, only slightly deviate from the semantic meaning of the queries. Examples of correct and incorrect translations are shown in Figure~\ref{fig:gpt_translation}. 
Based on these results, we conclude that GPT-4o is the most suitable LLM to implement the synthetic pair generation in \approach \textbf{(RQ1)}. 

In addition to measuring the accuracy of LLMs, we also assessed the quality of the translations provided in the \lcquad dataset.  
Table~\ref{tab:model-acc-manual} shows that these translations are only correct in 60\% of the cases. Most occurring errors here are semantic errors that do not align with the corresponding queries. 
Therefore, we have released a new version of the \lcquad dataset with synthetic translations generated and curated with \approach (cf. Section~\ref{sec:lcquadsynth}).

\begin{table}[t!]
  \centering
  \caption{Accuracy, precision, and recall of the verifier architectures with respect
  to the manual assessment of 300 queries of the \lcquad dataset.}
  \label{tab:filter-results}
\begin{tabular}{lcccccc}
    \toprule
    & \multicolumn{3}{c}{$BiEncoder_{Q, NL}$} & \multicolumn{3}{c}{$CrossEncoder_{Q, NL}$} \\
    \cmidrule(lr){2-4} \cmidrule(lr){5-7}
    \textbf{Model} & Acc & Prec & Rec & Acc & Prec & Rec \\
    \midrule
    Llama 3 7B   & 0.69 & 0.73 & 0.78 & 0.82 & 0.91 & 0.78 \\
    Gemini 2.0 Flash   & 0.87 & 0.91 & 0.95 & 0.91 & 0.94 & 0.97 \\
    o1-mini        & 0.89 & 0.95 & 0.93 & 0.92 & 0.97 & 0.95 \\
    GPT-4o    & 0.94 & 0.98 & 0.96 & 0.93 & 0.98 & 0.95 \\
    \lcquad        & 0.72 & 0.75 & 0.80 & 0.66 & 0.80 & 0.58 \\
    \midrule
        \textit{Average}       & 0.82 & 0.86 & \textbf{0.88} & \textbf{0.85} & \textbf{0.92} & 0.85 \\
    \bottomrule
  \end{tabular}
\end{table}

\begin{figure*}[h]
    \centering
    \includegraphics[width=\linewidth]{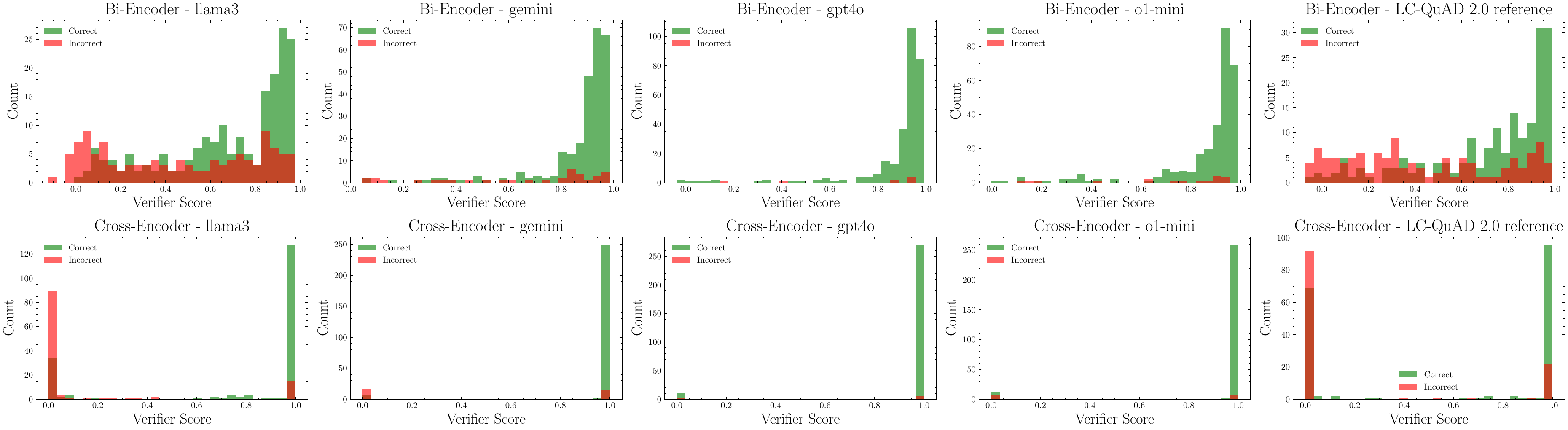} 
    \caption{Verifier predictions for correct and wrong translations across all models.}
    \label{fig:histogram}
\end{figure*}

\subsection{Results: Verifier Model Accuracy}
\label{sec:verifier_acc}
In this experiment, we investigate the performance of the verifier component using a bi-encoder ($BiEncoder_{Q, NL}$) or a cross-encoder ($CrossEncoder_{Q, NL}$) architecture. 
The verifier was trained using the synthetic pairs generated with GPT-4o. 
Next, we measure the performance of the verifier in distinguishing correct and wrong translations provided by the different studied models. 

We report on accuracy, precision, and recall at a threshold $\tau=0.5$. 
That is, if the verifier score \textit{Verifier$_{Q,NL}$} is greater than $0.5$, it is considered a correct translation; else, it is a wrong translation. 
Precision denotes how many wrong translations have actually been identified as wrong, while recall denotes how many correct translations have been identified as correct. 
As shown in Table~\ref{tab:filter-results}, despite the verifiers being trained on translations from GPT-4o, its performance generalizes to translations from other LLMs. More notably, it also performs well on the reference natural language translations from the \lcquad dataset, despite being significantly (and often grammatically wrong) different in paraphrasation than the translations of the LLMs. In terms of metrics, both the bi- and cross-encoder have comparable accuracy, where the bi-encoder has a higher recall while the cross-encoder has a higher precision. 

Figure~\ref{fig:histogram} shows the raw predictions of the two verifier architectures across all datasets in a histogram. The manual assessment of the corresponding queries is shown in red (wrong) and green (correct). The histograms show a stark difference in behavior between the models, where the bi-encoder predictions are more spread out, and the cross-encoder scores are concentrated around 0 and 1. This is expected since the bi-encoder uses cosine similarity between the embedded query and translation to assess its similarity, which is naturally a smoother metric, while the cross-encoder uses a small classifier on top of the encoder. This also implies that the bi-encoder can be more easily tuned for a higher precision/recall by changing the decision threshold (e.g., $\tau=0.5$ as used in Table~\ref{tab:filter-results}), while this would have no large effect on the cross-encoder. Generally, both models show that correctly translated queries are more heavily concentrated to the right (high confidence in correctness), while wrong translations are more evenly spread out and, especially for the cross-encoder, are concentrated to the left (low confidence in correctness). Hence, the ultimate choice between the verifier models depends on the specific requirements of the use case.

There are two predominant reasons for misclassifications of the verifier models. Firstly, since the LLMs receive detailed descriptions of the entities in their prompt (cf. Section~\ref{sec:q_to_nl}), they sometimes use alternative naming or paraphrasing based on the entity descriptions. Since this contrasts with the occurring labels in the SPARQL query, the verifier misclassifies the pair. Secondly, as explained in Section~\ref{sec:results_q_to_nl}, some LLMs frequently mistranslate queries involving reified statements. However, the verifier often assigns a high score. This can be mitigated by extending the input to the verifier to include descriptions for entities and predicates and by extending the synthetic translations used for training to focus more on correct and incorrect translations of queries involving blank nodes. An example of a misclassification of a query where the descriptions have been used rather than the entity names directly (cf.  Figure~\ref{fig:ex_verifier}). 

Based on these results, we conclude that the cross-encoder architecture is a more suitable architecture for \approach \textbf{(RQ2)}. In the following experiment, we will further investigate which of these architectures can improve the overall synthetic Q/NL dataset generated with the studied LLMs.

\begin{figure}
\begin{wrongbox_verifier}
    \textbf{Query} 
    \begin{lstlisting}[basicstyle=\ttfamily\small, frame=none, breaklines=true]
SELECT ?answer WHERE { 
 [figure of the Earth] [manifestation of] ?X . 
 ?X [opposite of] ?answer. }
\end{lstlisting}

    \textbf{Translation}\\
What is the opposite of the characteristic embodiment of the mathematical descriptions of Earth's complex shape?\\
\textbf{Verifier score}: 0.04
\end{wrongbox_verifier}
\caption{Misclassification of the verifier. The translation is considered a wrong translation due to the low score, yet, it is a correct translation. The translation includes verbalization from the descriptions instead of the entity labels that are occurring in the query.}
\label{fig:ex_verifier}
\end{figure}

\begin{table}[ht]
\caption{Manual accuracy before/after filtering ($\tau$=0.6) using the verifier and the fraction of original queries retained.}
\centering
\renewcommand{\arraystretch}{1.0} 
\setlength{\tabcolsep}{3.6pt} 
\begin{tabular}{lcccccc}
\toprule
\textbf{Model} & Llama3 & Gemini & o1-mini & GPT4o & \lcquad \\
\midrule
\multicolumn{6}{l}{\textbf{Bi-Encoder}} \\
\textit{before}    & 0.61 & 0.88 & 0.94 & 0.97 & 0.60 \\
\textit{after}     & \textbf{0.72} & \textbf{0.91} & \textbf{0.95} & \textbf{0.98} & \textbf{0.78} \\
\textit{retention} & 57\% & 90\% & 92\% & 93\% & 58\% \\
\midrule
\multicolumn{6}{l}{\textbf{Cross-Encoder}} \\
\textit{before}    & 0.61 & 0.88 & 0.94 & 0.97 & 0.60 \\
\textit{after}     & \textbf{0.91} & \textbf{0.93} & \textbf{0.97} & \textbf{0.98} & \textbf{0.81} \\
\textit{retention} & 52\% & 91\% & 92\% & 93\% & 43\% \\
\bottomrule
\end{tabular}
\label{tab:accuracy-filtering}
\end{table}

\begin{table*}[t!]
  \caption{Average values produced by the evaluation metrics when measuring the similarity between a SPARQL query and its generated NL translation. These pairs are obtained from 300 queries from the \lcquad dataset.}
  \label{tab:metrics}
  \centering
  \begin{tabular}{lcccccc}
    \toprule
    \textbf{Model} & $Acc_{manual}$ &\(\textit{Verifier}_{Q, NL}\) & \(BERTScore_{NL, GT}\) & Bleu & Rouge & Levenshtein \\
    \midrule
    Llama 3 7B  & 0.61 & 0.53 & 0.80 & 0.10  & 0.32 & 55.0 \\
    Gemini 2.0 Flash  & 0.88 & 0.91 & 0.85 & 0.16 & 0.41 & 42.0 \\
    o1-mini      &0.94 & 0.91 & 0.85 & 0.17 & 0.41 & 37.3 \\
    GPT-4o  & 0.97 & 0.94 & 0.85 & 0.17 & 0.41 & 39.5 \\
    \lcquad  &0.60 &  0.43 & -    & -    & -    & - \\
    \hline
    \(\rho(Acc_{manual}, \cdot)\) & --&  \(\textbf{1.00}\) & 0.77 & 0.95 & 0.77 & -0.80 \\
    \bottomrule
  \end{tabular}
\end{table*}

\subsection{Results: Accuracy Improvement with the Verifier}
We now investigate to which degree filtering of the dataset using the verifier models yields a quality improvement. For that, we filter out all queries from the 300 manually assessed queries that have a verifier score $< 0.6$. I.e., we use a threshold $\tau=0.6$ as the decision criterion to classify a Q/NL pair as correct or incorrect. Table~\ref{tab:accuracy-filtering} shows the manual accuracies of the queries before (same as Table~\ref{tab:model-acc-manual}) and after filtering. Additionally, it shows the \textit{retention} in percent, i.e., how many of the original 300 queries are retained after filtering. For all datasets, there is an improvement in the manual accuracy after filtering, with larger improvements for the datasets with worse initial quality (Llama 3 7B and \lcquad). Furthermore, the improvements are consistently greater for the cross-encoder model. Lastly, for all datasets, the retention rate is larger than 50\% except for \lcquad filtered using the cross-encoder verifier. Those results show that both verifier models can effectively be used as a data cleaning step for QA datasets, significantly enhancing the dataset's quality while retaining large parts of the dataset \textbf{(RQ3)}.

\subsection{Results: Verifier as a Metric}
\label{sec:metric}

In this study, we assess the effectiveness of the \textit{Verifier$_{Q,NL}$} score as a metric to measure the similarity between queries and translations. 
For this, we compare the metric scores with the accuracy values computed manually ($Acc_{manual}$, same as Table~\ref{tab:model-acc-manual}), and analyze how well the metrics reflect the results of a human-made evaluation. 

Table~\ref{tab:metrics} shows the values of the NLP metrics described in Section~\ref{sec:experimental_setup} alongside the average score of  \approach implemented with the cross-encoder architecture. 
The average verifier score is close to the accuracy of the manually assessed translation for most datasets, followed by $BERTScore_{NL,GT}$. The other NLP metrics have significantly different values from our reference ($Acc_{manual}$). 

Furthermore, we also report the rank correlation $\rho$ between the manually determined accuracy $Acc_{manual}$ and the different metrics. 
Rank correlation measures the degree of similarity between the rankings of two variables, in that case, how well one metric preserves the relative ordering of accuracy values compared to manual evaluation. 
The last row in Table~\ref{tab:metrics} shows that the verifier has the highest and a perfect rank correlation of 1.0, i.e., the ordering agrees exactly with the manual assessment. This shows that the verifier can be used as an improved drop-in metric for other NLP metrics for sets of translations between SPARQL and natural language \textbf{(RQ4)}. Additionally, unlike all other compared metrics, the verifier does not require a reference (or gold standard) translation, it just requires the query and its translation.
This makes our approach applicable even when there is no access to gold standard data. 

\subsection{Results: Applications to NL $\rightarrow$ Q Translation}
\label{sec:nl_to_q}

Lastly, we investigate if the synthetic pairs generated by \approach can improve an LLM in translating NL questions to SPARQL queries, which is a core KGQA task. 
In this evaluation, we deployed \approach using the best  configurations reported in previous experiments: 
(i) we use GPT-4o to generate synthetic pairs of SPARQL query and translation as training data for the verifier,  and 
(ii) we use the cross-encoder transformer architecture for the verifier. 

Next, we compare the performance of three models based on GPT-4o-mini using data generated by \approach. 
\textit{Few Shot} is a GPT-4o-mini model without finetuning, i.e., only by providing the prompt using example translations and entity/predicate descriptions. 
\textit{Synthetic} is the GPT-4o-mini model trained using $500$ randomly drawn pairs synthetically generated with the Q $\rightarrow$ NL generator of \approach.
\textit{Synthetic Filtered} is the GPT-4o-mini model trained using $500$ randomly draw synthetic pairs generated and verified by \approach with a score $> 0.5$. 
We trained the GPT-4o-mini \textit{Synthetic} and \textit{Synthetic Filtered}  models using the OpenAI API, which costs approximately \$2 per model. 

Table~\ref{tab:nl_to_q_results} shows the accuracy of the different LLM models.  
The untrained \textit{Few Shot} GPT-4o-mini  is only able to translate the NL to a query correctly in 43\% of cases, while the finetuned model (\textit{Synthetic}) shows a stark improvement to 88\%. 
This shows that the synthetically generated pairs with \approach already constitute an effective training dataset for QA systems. 
Lastly, filtering the synthetic dataset using the verifier model before training leads to a further increase in translation accuracy to 89\%.  
Considering that the GPT family of LLMs are considered frontier LLMs with outstanding results in many NL tasks, our approach is still able to improve its performance. 
These results show that \approach can improve KGQA systems \textbf{(RQ5)} by providing cleaner training data.

\begin{table}[h!]
  \centering
  \caption{Accuracy of NL to SPARQL query translation of GPT-4o-mini models. \textit{Few Shot} is an untrained model; \textit{Synthetic} is fine-tuned on Q $\rightarrow$ NL  translations with \approach; \textit{Filtered Synthetic} fine-tuned with generated and verified translations with \approach.}
  \label{tab:nl_to_q_results}
  \begin{tabular}{lcccc}
    \cmidrule(lr){2-5}
    & \textbf{Few Shot} 
    & \textbf{Synthetic} 
    & \textbf{Synthetic Filtered} \\
    \midrule
    \textbf{$\textrm{Accuracy}$} & 0.43& 0.88 & \textbf{0.89} & \\
    \bottomrule
  \end{tabular}
\end{table}

\section{LC-QuAD 2.0-synth: Benchmark Extension with \approach}\label{sec:lcquadsynth}
To further improve the quality of benchmark datasets for QA systems, we provide an updated version of the \lcquad dataset. In this version, we supplement each query with an additional LLM-generated natural language translation of the query alongside the verifier scores.

\smallbreak
\noindent
\textbf{Dataset Availability.} 
The full dataset, including the newly generated translations and verifier scores, is publicly available on Hugging Face\footnote{\url{https://huggingface.co/datasets/timschwa/lc_quad_synth}}, 
one of the most popular repositories for QA datasets. 

\smallbreak
\noindent
\textbf{Dataset Extension.}
For each query in \lcquad, we generate a corresponding natural language question using GPT-4o, leveraging the prompt-based approach described in Section~\ref{sec:q_nl_generation}. These synthetic translations provide a high-quality alternative to human-generated translations with better semantic similarity (as determined using our verifier model). The updated dataset includes
\begin{enumerate}
    \item Original \lcquad data: Including the original wikidata and DBpedia queries as well as the human-generated translations.
    \item The wikidata queries with IRIs replaced by their corresponding labels.
    \item The descriptions of all occurring entities and predicates in the wikidata query.
    \item LLM-generated translations: Generated using GPT-4o, following the few-shot prompting methodology outlined in our approach.
    \item Verifier scores: For both human and LLM translations, assessing their semantic consistency with the original SPARQL queries.
\end{enumerate}
\noindent
\textbf{Comparison of Human vs. LLM Translations}
To evaluate the quality of the newly provided translations, we compute the average verifier score (using the cross-encoder model) for both human-generated and LLM-generated translations.
Our results indicate that LLM-generated translations achieve a significantly higher average verifier score (0.93) compared to human-generated translations (0.46). This suggests that the LLM-based translations provide more semantically consistent translations across the whole scale of the dataset. In summary, the updated \lcquad constitutes a new reliable dataset for training and evaluating QA systems. 
\section{Conclusion}
In this work, we have demonstrated that large language models can translate structured queries (SPARQL) into natural language with high accuracy, even without additional fine-tuning. This capability enables the generation of high-quality synthetic query-translation pairs, which can be leveraged for training verifier models that assess the correctness of query to natural-language translations. Our  approach \approach generalizes across different sources of translations, including other LLMs and human-generated translations, and outperforms existing NLP metrics in evaluating the quality of query-natural language pairs.

Beyond evaluating translations, we showed that incorporating \approach into Question Answering (QA) pipelines improves natural language to query translation. This suggests that the verifier has multiple practical applications, such as providing live feedback in query-based tools, serving as a loss function in training QA systems, and refining datasets used for QA tasks. By enabling more robust and semantically accurate query translations, our verifier framework has the potential to enhance the usability and effectiveness of knowledge graph-based QA systems. We further release a new version of the \lcquad dataset including our improved LLM-based translations that can be used both for building and evaluating QA systems.

\paragraph{Limitations and Future Work} Despite these promising results, our approach has certain limitations. First, our experiments have been conducted on a limited dataset (\lcquad), and additional evaluations across diverse domains, query types, and datasets are necessary to confirm broader generalizability. Additionally, while LLMs perform well in query-to-natural-language translation, fine-tuning them specifically for this task could further improve performance and consistency. Further, the verifier model is currently trained on a relatively small set of 6,000 queries. Expanding the training data with more diverse queries and synthetic examples could enhance its robustness. Moreover, incorporating descriptions of entities and predicates into the verifier model may help reduce misclassifications caused by paraphrasing and entity aliasing. Lastly, improving the diversity of synthetic translations by introducing more targeted perturbations reflecting the verifier misclassifications will further improve its accuracy.

\bibliographystyle{plain}  
\bibliography{references}

\end{document}